\documentclass{llncs}
\usepackage{import}
\usepackage{ifthen}
\usepackage{url}
\PassOptionsToPackage{dvipsnames}{xcolor}
\usepackage{xcolor}
\usepackage{amssymb}
\usepackage{amsmath}
\usepackage{import}
\usepackage{xparse}
\usepackage{amsmath}
\usepackage{xspace}
\usepackage{datatool}
\usepackage{glossaries}
\providecommand\m[1]{\ensuremath{#1}\xspace}
\renewcommand{\m}[1]{\ensuremath{#1}\xspace}

	\newcommand{\lrule}{\leftarrow}
	\newcommand{\cause}{\stackrel{c}{\lrule}}

	\newcommand{\voc}{\m{\Sigma}}

	\newcommand{\struct}{\m{I}}

	\newcommand{\theory}{\m{\mathcal{T}}}

	\NewDocumentCommand\inter{g+g}{%
	  \IfNoValueTF{#1}
	    {\struct}
	    {\m{#1^{#2}}}}

	\renewcommand{\int}{\m{\mathbb{Z}}}

	\NewDocumentCommand\subs{g+g}{%
	  \IfNoValueTF{#1}
	    {\m{/}}
	    {\m{#1/ #2}}}

	\newcommand{\logicname}[1]{\textsc{#1}\xspace}

	\newcommand{\idp}{\logicname{IDP}}

	\newcommand{\minisatid}{\logicname{MiniSAT(ID)}}

	\newcommand{\fodot}{\logicname{FO(\ensuremath{\cdot})}}

\newcommand{\ouracronym}[3]{%
	\newacronym{#1}{#2}{#3}
	\expandafter\newcommand\csname #1\endcsname{\gls{#1}\xspace}%
}
	\ouracronym{FO}{FO}{first-order logic}
	\ouracronym{PC}{PC}{propositional calculus}
	\ouracronym{MX}{MX}{Model Expansion}
	\ouracronym{MO}{MO}{Model Optimization}
	\ouracronym{ASP}{ASP}{Answer Set Programming}
	\ouracronym{TP}{TP}{Theorem Proving}
	\ouracronym{LP}{LP}{Logic Programming}
	\ouracronym{CP}{CP}{Constraint Programming}
	\ouracronym{FP}{FP}{Functional Programming}
	\ouracronym{KR}{KR}{Knowledge Representation}
	\ouracronym{CSP}{CSP}{Constraint Satisfaction Problem}
	\ouracronym{SMT}{SMT}{SAT Modulo Theories}
	\ouracronym{KBS}{KBS}{knowledge base system}
	\ouracronym{NNF}{NNF}{Negation Normal Form}
	\ouracronym{FNNF}{FNNF}{Flat Negation Normal Form}
	\ouracronym{DefNNF}{DefNNF}{Definition Negation Normal Form}
	\ouracronym{DEFNF}{DEFNF}{Definition Normal Form}
	\ouracronym{CDCL}{CDCL}{Conflict-Driven Clause-Learning}
	\ouracronym{WFS}{WFS}{Well-Founded Semantics}
	\ouracronym{LCG}{LCG}{Lazy Clause Generation}
	\ouracronym{AEL}{AEL}{Autoepistemic Logic}
	\ouracronym{OEL}{OEL}{Ordered Epistemic Logic}
	\ouracronym{AFT}{AFT}{Approximation Fixpoint Theory}

	\makeatletter
	\def\ifenv#1{
	\def\@tempa{#1}%
	\def\@ttempa{#1*}%
	\ifx\@tempa\@currenvir
	\expandafter\@firstoftwo
	\else
	\expandafter\@secondoftwo
	\fi
	}
	\makeatother

	\newcommand{\ddrule}[4]{\ensuremath{#1 \leftarrow #2 & \{#3\} & #4}}
	\newcommand{\drule}[2]{\ensuremath{#1 & \leftarrow & #2}}

	\newcommand{\darule}[4]{\ensuremath{#1 \leftarrow #2 & \{#3\} & #4}}
	\newcommand{\arule}[2]{\ensuremath{#1 \, &\leftarrow \, #2}}

	\newcommand{\LNDRule}[2]{
	\ifenv{array}
	{\drule{#1}{#2}}
	{ \ifenv{align}
		{\arule{#1}{#2}}
		{\ifenv{align*}
		{\arule{#1}{#2}}
		{ERROR: using LDRule in unsupported environment: \@currenvir}
		}
	}
	}

	\newcommand{\LDRule}[4]{
	\ifenv{array}
	{\ddrule{#1}{#2}{#3}{#4}}
	{ \ifenv{align}
		{\darule{#1}{#2}{#3}{#4}}
		{\ifenv{align*}
		{\darule{#1}{#2}{#3}{#4}}
		{ERROR: using LDRule in unsupported environment: \@currenvir}
		}
	}
	}

	\NewDocumentCommand\LRule{m+g+g+g}{%
		\IfNoValueTF{#2}%
		{#1.&}{%
		\IfNoValueTF{#3}
		{\LNDRule{#1}{#2.}}
		{\LDRule{#1}{#2.}{#3}{#4}}%
		}
	}

	\NewDocumentCommand\CLRule{m+g}{%
	\ifenv{array}
	{\cdrule{#1}{#2}}
	{ \ifenv{align}
		{\carule{#1}{#2}}
		{\ifenv{align*}
			{\carule{#1}{#2}}
			{ERROR: using CLRule in unsupported environment: \@currenvir}
		}
	}
	}

	\NewDocumentCommand\carule{m+g}{%
		\IfNoValueTF{#2}
			{\ensuremath{#1.}}
			{\ensuremath{#1 \, &\cause \, #2}}}
	\NewDocumentCommand\cdrule{m+g}{%
		\IfNoValueTF{#2}
			{\ensuremath{#1.}}
			{\ensuremath{#1 & \cause & #2}}}

	\newcommand{\algrule}[4]{
	\hbox{{#1}:}& 
	\quad #2 ~\longrightarrow~ #3 
	\hbox{~ if } #4\\
	}

	\newcommand{\AlgoRule}[4]{
	\ifenv{array}
	{\algrule{#1}{#2}{#3}{#4}}
		{ERROR: using AlgoRule in unsupported environment: \@currenvir}
	}

	\newcommand{\ignore}[1]{}

	\newboolean{nocomments}
	\setboolean{nocomments}{false}

	\newboolean{commentmargin}
	\setboolean{commentmargin}{true}

	\newcommand{\namedcomment}[3]{%
		\ifthenelse{\boolean{nocomments}}%
		{}%IF no comments, write nothing
		{%Otherwise
			\ifthenelse{\boolean{commentmargin}}%
				{ {\color{#3} \marginpar{\color{#3}\sc #2}#1}  }%Name in margin
				{  {\color{#3} {\sc #2}: #1}  }%Name not in margin
		}%
	}
	\newcommand{\mnamedcomment}[3]{\ifthenelse{\boolean{nocomments}}{}{{\marginpar{\tiny \color{#3}{\sc #2}:#1}}}}

	\usepackage{soul}

\usepackage[normalem]{ulem} % wavy underlines
\makeatletter
\font\uwavefont=lasyb10 scaled 700
\def\spelling{\bgroup\markoverwith{\lower3.5\p@\hbox{\uwavefont\textcolor{Red}{\char58}}}\ULon}
\def\grammar{\bgroup\markoverwith{\lower3.5\p@\hbox{\uwavefont\textcolor{LimeGreen}{\char58}}}\ULon}
\def\phrasing{\bgroup\markoverwith{\lower3.5\p@\hbox{\uwavefont\textcolor{RoyalBlue}{\char58}}}\ULon}

\newcommand\remove{\bgroup\markoverwith{\textcolor{red}{\rule[0.5ex]{2pt}{0.4pt}}}\ULon}
\makeatother

\usepackage{etoolbox}
\ExplSyntaxOn
\newcommand\setcitation[2]{%
	\csdef{mycommoncitation\text_uppercase:n{#1}}{#2}}
\newcommand\getcitation[1]{%
	\csuse{mycommoncitation\text_uppercase:n{#1}}}

\ExplSyntaxOff

\setcitation{IDP}{WarrenBook/DeCatBBD18}
\setcitation{fodot}{tocl/DeneckerT08}
\setcitation{CPsupport}{ictai/DeCat13}
\setcitation{minisatid}{ictai/DeCat13}
\setcitation{FunctionDetection}{iclp/DeCatB13}
\setcitation{lazyGrounding}{jair/CatDBS15} 
\setcitation{Bootstrapping}{ngc/BogaertsJDJBD16}
\setcitation{BootstrappingIDP}{ngc/BogaertsJDJBD16}
\setcitation{FOSymmetry}{tplp/DevriendtBBD16}
\setcitation{IDP2}{lash08/WittocxMD08}
\setcitation{FOLASP}{tplp/VanDesselDV21}

\setcitation{foid}{tocl/DeneckerT08}
\setcitation{cplogic}{journal/tplp/VennekensDB10}

\setcitation{fodot2asp}{corr/DeneckerLTV19} %TODO replace by journal publication if one is publishedr
\setcitation{Tarskian}{corr/DeneckerLTV19} %Same as the one above 
\setcitation{TarskianSemanticsASP}{corr/DeneckerLTV19} %Same as the one above 

\setcitation{SAT}{faia/336}
\setcitation{HandbookOfSAT}{faia/336}

\setcitation{DPLLT}{cav/GanzingerHNOT04}
\setcitation{SMT}{faia/BarrettSST21}

\setcitation{CP}{fai/Rossi06}
\setcitation{Inca}{iclp/DrescherW12}
\setcitation{csp2asp}{ijcai/DrescherW11}
\setcitation{EZCSP}{lpnmr/Balduccini11}

\setcitation{ASPComp2}{lpnmr/DeneckerVBGT09}
\setcitation{ASPComp3}{journals/tplp/CalimeriIR14}
\setcitation{ASPComp4}{conf/lpnmr/AlvianoCCDDIKKOPPRRSSSWX13}
\setcitation{ASPComp5}{journals/ai/CalimeriGMR16}
\setcitation{ASPComp6}{jair/GebserMR17}
\setcitation{ASPComp7}{tplp/GebserMR20}
\setcitation{AspInPractice}{synthesis/2012Gebser}
\setcitation{clasp}{ai/GebserKS12}
\setcitation{oclingo}{kr/GebserGKOSS12}
\setcitation{clingo}{iclp/GebserKKOSW16}
\setcitation{gringo}{lpnmr/GebserST07}
\setcitation{cmodels}{aaai/GiunchigliaLM04}
\setcitation{DLV}{tocl/LeonePFEGPS06}
\setcitation{GroundingBottleneck}{padl/SonPL14}

\setcitation{MaxSAT}{faia/LiM21}

\setcitation{KR}{Baral:2003}

\setcitation{inputster}{tplp/Jansen13}
\setcitation{LearningPaper}{TPLP/BruynoogheBBDDJLRDV} 
\setcitation{clog}{iclp/BogaertsVDV14} %TODO replace by journal publication if one is published
\setcitation{foc}{iclp/BogaertsVDV14}
\setcitation{inferenceClog}{ecai/BogaertsVDV14}
\setcitation{inferenceFOC}{ecai/BogaertsVDV14}
\setcitation{examplesClog}{nmr/BogaertsVDV14b} %TODO replace by better publication if possible
\setcitation{AFT}{DeneckerMT00}
\setcitation{KBS}{iclp/DeneckerV08}
\setcitation{KBS-invitedtalk}{jelia/Denecker16}
\setcitation{KBPE}{inap/DePooterWD11}
\setcitation{lazygroundingASP}{ijcai/BogaertsW18} 
\setcitation{justifications}{lpnmr/DeneckerBS15} 
\setcitation{justificationTheory}{lpnmr/DeneckerBS15} 
\setcitation{justificationsAlpha}{ijcai/BogaertsW18} 
\setcitation{ASP}{marek99stable} %{marek99stable,Niemela99,iclp/Lifschitz99} 
\setcitation{satid}{sat/MarienWDB08}
\setcitation{lazyclausegeneration}{constraints/OhrimenkoSC09}
\setcitation{FP}{ACMCS/Hudak89}
\setcitation{GroundingWithBounds}{jair/WittocxMD10}
\setcitation{GroundWithBounds}{jair/WittocxMD10}

\setcitation{LTC}{iclp/Bogaerts14}
\setcitation{SPSAT}{ictai/DevriendtBMDD12}
\setcitation{BreakID}{sat/DevriendtBBD16}
\setcitation{LCG}{stuckeyLCG}
\setcitation{MiniZinc}{conf/cp/NethercoteSBBDT07}
\setcitation{GroundedFixpoints}{ai/BogaertsVD15}
\setcitation{PartialGroundedFixpoints}{ijcai/BogaertsVD15}
\setcitation{LogicBlox}{datalog/GreenAK12}
\setcitation{proB}{journals/sttt/LeuschelB08}
\setcitation{NaturalInductions}{KR/DeneckerV14} %TODO replace by journal publication if one is published
\setcitation{LP}{jacm/EmdenK76}

\setcitation{AF}{ai/Dung95}
\setcitation{ADF}{kr/BrewkaW10}
\setcitation{af}{ai/Dung95}
\setcitation{adf}{kr/BrewkaW10}
\setcitation{ADFRevisited}{ijcai/BrewkaSEWW13}
\setcitation{DefaultLogic}{ai/Reiter80}
\setcitation{DL}{ai/Reiter80}
\setcitation{AEL}{mo85}
\setcitation{minisat}{sat/EenS03}
\setcitation{completion}{adbt/Clark78}
\setcitation{ClarkCompletion}{adbt/Clark78}
\setcitation{wasp}{lpnmr/AlvianoDFLR13}
\setcitation{lcg}{stuckeyLCG}
\setcitation{CEGAR}{jacm/ClarkeGJLV03}
\setcitation{CuttingPlane}{or/DantzigFJ54}
\setcitation{CuttingPlanes}{or/DantzigFJ54}
\setcitation{kodkod}{tacas/TorlakJ07}
\setcitation{CDCL}{Marques-SilvaS99}
\setcitation{1UIP}{iccad/ZhangMMM01}
\setcitation{relevance}{ijcai/JansenBDJD16}
\setcitation{relevance-implementation}{aspocp/JansenBDJD16}
\setcitation{WFS}{GelderRS91}
\setcitation{UnfoundedSet}{GelderRS91}
\setcitation{UFS}{GelderRS91}
\setcitation{stablesemantics}{iclp/GelfondL88}
\setcitation{shatter}{Shatter}
\setcitation{sbass}{drtiwa11a}
\setcitation{lparsemanual}{url:lparsemanual}
\setcitation{AIC}{ppdp/FlescaGZ04}
\setcitation{templates}{tplp/DassevilleHJD15}
\setcitation{templates2}{iclp/DassevilleHBJD16}
\setcitation{sat-to-sat}{aaai/JanhunenTT16}
\setcitation{sat-to-sat-qbf}{bnp/BogaertsJT16}
\setcitation{sat-to-sat-SO}{kr/BogaertsJT16}
\setcitation{XSB}{SwiW12}
\setcitation{KCmap}{jair/DarwicheM02}
\setcitation{KnowledgeCompilationMap}{jair/DarwicheM02}
\setcitation{TLA}{DBLP:books/aw/Lamport2002}
\setcitation{EventB}{BookAbrial2010}
\setcitation{MX}{MitchellT05}
\setcitation{MIP}{Sierksma96}
\setcitation{PerfectModel}{minker88/Przymusinski88}
\setcitation{SafeInductions}{ijcai/BogaertsVD17}
\setcitation{AIC}{ppdp/FlescaGZ04}
\setcitation{alpha}{lpnmr/Weinzierl17}
\setcitation{omiga}{jelia/Dao-TranEFWW12}
\setcitation{gasp}{fuin/PaluDPR09}
\setcitation{asperix}{lpnmr/LefevreN09a}
\setcitation{CTL}{lop/ClarkeE81}
\setcitation{AFT-AIC}{ai/BogaertsC18}
\setcitation{UltimateApproximator}{DeneckerMT04}
\setcitation{KripkeKleene}{Fitting85}
\setcitation{AFT-HO}{corr/CharalambidisRS18} %TODO replace
\setcitation{HereThere}{Heyting30}
\setcitation{dAEL}{ijcai/HertumCBD16}
\setcitation{SDD}{ijcai/Darwiche11}
\setcitation{HEX}{ijcai/EiterIST05}
\setcitation{wADF}{aaai/BrewkaSWW18}
\setcitation{wADFfix}{corr/BrewkaSWW18}
\setcitation{TransitionSystems}{jacm/NieuwenhuisOT06}
\setcitation{galliwasp}{lopstr/MarpleG12}
\setcitation{GalliWasp}{lopstr/MarpleG12}
\setcitation{clingcon}{tplp/BanbaraKOS17}
\setcitation{lp2sat}{birthday/JanhunenN11}
\setcitation{lp2mip}{LIU12}
\setcitation{lp2diff}{lpnmr/JanhunenNS09}
\setcitation{lp2acyc}{ecai/GebserJR14}
\setcitation{PB}{faia/RousselM09}
\setcitation{pb}{faia/RousselM09}
\setcitation{CuttingPlanes}{dam/CookCT87}
\setcitation{RoundingSAT}{ijcai/ElffersN18}
\setcitation{PRS}{aaai/DixonG02}
\setcitation{sat4j}{jsat/BerreP10}
\setcitation{mingo}{LIU12}
\setcitation{pbmodels}{lpnmr/LiuT05}
\setcitation{HEF-LP}{lpnmr/GebserLL07}
\setcitation{HCF-LP}{amai/Ben-EliyahuD94}
\setcitation{aspcore2}{tplp/CalimeriFGIKKLM20}
\setcitation{AspCore2}{tplp/CalimeriFGIKKLM20}
\setcitation{SemanticWeb}{book/AntoniouGHH12}
\setcitation{QBF}{faia/BuningB09}
\setcitation{SMUS}{cp/IgnatievPLM15}
\setcitation{CDCLsym}{tacas/MetinBCK18}
\setcitation{SEL}{sat/DevriendtBB17}
\setcitation{Coq}{txtcs/BertotC04}
\setcitation{Isabelle}{book/NipkowPW02}
\setcitation{HOL}{tphol/SlindN08}
\setcitation{lean}{cade/MouraU21}
\setcitation{CakeML}{jfp/TanMKFON19}
\setcitation{FRAT}{lmcs/BaekCH22}
\setcitation{DRUP}{date/GoldbergN03}
\setcitation{RUP}{date/GoldbergN03}
\setcitation{GRIT}{tacas/Cruz-FilipeMS17}
\setcitation{TraceCheck}{url:TraceCheck}
\setcitation{LRAT}{cade/Cruz-FilipeHHKS17}
\setcitation{PR}{cade/HeuleKB17}
\setcitation{DRAT}{sat/WetzlerHH14}
\setcitation{satcomp2016}{aaai/BalyoHJ17}
\setcitation{satcomp2013}{url:SATcomp2013}
\setcitation{Zinc}{constraints/MarriottNRSBW08}
\setcitation{MiniZinc}{cp/NethercoteSBBDT07}
\setcitation{Essence}{constraints/FrischHJHM08}
\setcitation{smodels}{lpnmr/SyrjanenN01}
\setcitation{lparse}{Syrjanen98}
\setcitation{IDPZ3}{corr/CarbonelleVVD22} %TODO replace by real publication
\setcitation{IDP-Z3}{corr/CarbonelleVVD22} %TODO replace by real publication
\setcitation{Z3}{tacas/MouraB08}

\setcitation{DMN}{DMN} %TODO is there something better than just this on-line thing?
\setcitation{Planning}{book/GhallabNT04}
\setcitation{AutomatedPlanning}{book/GhallabNT04}
\setcitation{RC2}{jsat/IgnatievMM19}
\setcitation{enfragmo}{phd/Aavani14}
\setcitation{}{}
  
 \ExplSyntaxOn
\newcommand\refto[1]{%
      \ifcsname  mycommoncitation\text_uppercase:n{#1}\endcsname%
      \getcitation{#1}%
      \else%
      #1%
      \fi%
      }

\newcommand\mycite[1]{%
      \ifcsname mycommoncitation\text_uppercase:n{#1}\endcsname%
   \cite{\getcitation{#1}}%
  \else%
    \cite{#1}%
  \fi%
}	
  
\newcommand\mycitet[1]{%
      \ifcsname mycommoncitation\text_uppercase:n{#1}\endcsname%
   \citet{\getcitation{#1}}%
  \else%
    \citet{#1}
  \fi%
}	
  
\ExplSyntaxOff
\ignore{

}

\usepackage[]{algorithm2e}
\usepackage{comment}
\usepackage{mdwlist}

\newcommand{\CSOP}{COP}
\newcommand{\LMOP}{MOP}
\newcommand{\idpsym}{DES}
\newcommand{\structt}{\m{J}}
\newcommand{\solspace}{Sol}
\newcommand{\assspace}{As}

\begin{document}

\linespread{0.94}

\begin{comment}
\setlength{\parskip}{0pt}
\setlength{\parsep}{0pt}
\setlength{\headsep}{0pt}
\setlength{\topskip}{0pt}
\setlength{\topmargin}{0pt}
\setlength{\topsep}{0pt}
\setlength{\partopsep}{0pt}
\end{comment}

\title{Transforming Constraint Programs to Input for Local Search}

%If you're using runningheads you can add an abreviated title for the running head on odd pages using the following
%\titlerunning{abreviated title goes here}
%and an alternative title for the table of contents:
%\toctitle{table of contents title}

\subtitle{On the relation between symmetry and neighborhoods}

%For a single author
%\author{Author Name}

%For multiple authors:
\author{Jo Devriendt \and Patrick De Causmaecker \and Marc Denecker}

%If using runnningheads you can abbreviate the author name on even pages:
%\authorrunning{abbreviated author name}
%and you can change the author name in the table of contents
%\tocauthor{enhanced author name}

%For a single institute
%\institute{Institute Name \email{email address}}
% If authors are from different institutes 
\institute{University of Leuven\\ \email{\{jo.devriendt,patrick.decausmaecker,marc.denecker\}@cs.kuleuven.be}}

%to remove your email just remove '\email{email address}'
% you can also remove the thanks footnote by removing '\thanks{Thank you to...}'

\maketitle

\begin{abstract}
Applying local search algorithms to combinatorial optimization problems is not an easy feat. Typically, human intervention is required to compile the constraints to input data for some metaheuristic algorithm. In this paper, we establish a link between symmetry properties of constraint optimization problems and local search neighborhoods, and we use this link to automatically generate neighborhoods from a constraint specification in the context of the \idp system. We evaluate the obtained neighborhoods for six classical optimization problems. The resulting observations support the viability of this technique.
\end{abstract}

\section{Introduction}
Combinatorial optimization problems are studied across many branches of science, and are abundant in industrial applications and real-life scenarios. One way to solve combinatorial optimization problems is to generate an initial, suboptimal solution, and to iteratively refine the solution until a stop-criterion is met and a hopefully optimal solution is achieved. This is roughly the process studied by the Metaheuristics community, which investigates techniques such as genetic algorithms, local search, hyperheuristics, swarm-based optimization etc. These techniques have proven very effective for many problems, especially for large problems with many (suboptimal) solutions.

The field of Constraint Programming (CP) also aims to solve combinatorial optimization problems effectively, but most CP systems perform a complete traversal of the search space by use of search trees with propagation, satisfiability solving, and/or mixed integer programming. These complete approaches are often effective, but potentially limit the size of the optimization problems that can efficiently be tackled.

In this paper, we investigate how we can bring both fields of research closer together by providing an automated way of transforming a constraint optimization problem specification into input for local search algorithms. The method is based on symmetry properties of optimization problems that allow to transform suboptimal solutions into hopefully better solutions.

This paper is organized in the following way. In Section \ref{cp_ls}, we give an abstract description of what a constraint optimization problem is, how it relates to certain forms of local search, and how symmetry can be used to derive input neighborhoods for local search algorithms from a constraint optimization specification. The next section describes in more detail how this is done for a constraint optimization specification for the \idp. Section \ref{sec_experiments} reports on the nature of local search neighborhoods derived by the \idp system on multiple constraint optimization specifications. After subsequently sketching related work, the paper concludes.

\section{Constraint Programming and Local Search}
\label{cp_ls}
\subsection{Constraint Satisfaction Optimization Problems}
A \CSOP{} can be characterized as a quadruple $(V,D,C,O)$ where $V$ denotes a set of variables, $D$ a domain of possible values for these variables, $C$ a set of constraints and $O$ an objective function. An \emph{assignment} to a CSP $\Pi=(V,D,C,O)$ is a function $\alpha: V \rightarrow D$. We refer to the set of all assignments to a \CSOP{} $\Pi$ as its \emph{assignment space} $\assspace_\Pi$.

We abstract a constraint $c \in C$ as the subset of $\assspace_\Pi$ for which $c$ is satisfied. So, $\alpha \in c$ means $\alpha$ satisfies $c$, and $\alpha \not \in c$ means $\alpha$ violates $c$. A \emph{satisfying assignment} to a \CSOP{} is an assignment which satisfies all constraints in $C$. We refer to the set of all satisfying assignments of a \CSOP{} $\Pi$ as the \emph{solution space} $\solspace_\Pi$. Note that the solution space is the intersection of all constraints -- $\solspace_\Pi=\cap_{i} c_i$ with $c_i \in C$ -- and that a \CSOP{} is unsatisfiable if the solution space is empty.

An objective function $O:(V \rightarrow D)\rightarrow \mathbb{N}$ maps assignments to natural numbers. Without loss of generality, we take an \emph{optimal solution} to a \CSOP{} to be a satisfying assignment that is minimal with respect to the objective function. More formally, a satisfying assignment $\alpha \in \solspace_\Pi$ is an optimal solution if $\forall \alpha' \in \solspace_\Pi: O(\alpha) \leq O(\alpha')$.

%\bart{The field of Constraint Programming is concerned with finding solutions of...}

\begin{example}
\label{ex_TSP}
A classical \CSOP{} is the Traveling Salesman Problem (TSP). We can model this problem as a set of cities that need to be visited in a certain order, in such a way that the total distance of the visited tour is minimal. Given $n$ cities, we can use $V=\{v_0,\ldots,v_{n-1}\}$ as a set of variables, the set of cities as domain $D$, and $C$ containing the singular constraint that all variables must be assigned a different city. Given a distance matrix between cities $Dist:D\times D \rightarrow \mathbb{N}$, the objective function $O: \alpha \mapsto \sum_i Dist(\alpha(v_i),\alpha(v_{(i+1)\mod n}))$ maps each assignment to the sum of the distances between two subsequent cities.
\end{example}

\subsection{Local search algorithm}
Local search algorithms use the concept of a \emph{neighborhood} to perform a heuristic walk through the solution space of a \CSOP{}. 
\begin{definition}
\label{def_neighborhoods}
A \emph{neighborhood} $N$ for a \CSOP{} $\Pi$ is a mapping of each satisfying assignment to a set of satisfying assignments $N:\solspace_\Pi \rightarrow \mathcal{P}(\solspace_\Pi)$. $N(\alpha)$ is referred to as the set of \emph{neighbors} of a satisfying assignment $\alpha$ under $N$.
\end{definition}

Local search approaches such as those based on simulated annealing or tabu search require as input a neighborhood $N$ and some initial satisfying assignment $\alpha$. Given these, a typical local search algorithm explores the solution space by enumerating the neighbors of $\alpha$ under $N$. When some neighbor $\alpha' \in N(\alpha)$ satisfies an acceptance criterion (typically based on the objective value of $\alpha'$), it is accepted and becomes the new focus of attention. In a sense, the local search algorithm \emph{moves} from $\alpha$ to $\alpha'$, which is also expressed as executing a \emph{move}.

Search continues by exploring the neighbors of $\alpha'$, until a new neighbor is accepted, leading to a new move, repeating the loop. This loop ends when some stop criterion is met, and the satisfying assignment with the lowest objective value encountered during the search is returned.

The above notion of local search is in a way restrictive, since it does not capture optimization approaches such as evolutionary programming or swarm-based optimization. Nonetheless, metaheuristic methods such as tabu search, simulated annealing, variable neighborhood search or greedy optimization can be characterized by moving from satisfying assignment to satisfying assignment using neighborhoods.

\begin{example}
\label{ex_TSP_neighborhood}
For the TSP problem, a typical neighborhood is the so-called \emph{2-opt} neighborhood. This neighborhood maps each TSP-tour $\alpha$ to a set of new tours by removing a pair of edges, say between cities $c_1,c_2$ and $c_3,c_4$, and reconnecting the resulting two TSP-subpaths by introducing an edge between $c_1,c_4$ and an edge between $c_2,c_3$.
\end{example}

\subsection{Symmetry}
Given the above definition of a neighborhood, it is interesting to investigate its relationship with the notion of symmetry.
We follow \cite{Cohen06symmetrydefinitions} and define a symmetry of a $\CSOP{}$ as a permutation on the assignment space which preserves satisfaction to the constraints:
\begin{definition}
\label{def_symmetry}
A \emph{symmetry} $S$ for a \CSOP{} $\Pi=(V,D,C,O)$ is a permutation on the assignment space $S:(V \rightarrow D) \rightarrow (V \rightarrow D)$ such that $\alpha \in \solspace_\Pi \Leftrightarrow S(\alpha) \in \solspace_\Pi$.
\end{definition}

By Definition \ref{def_symmetry}, every permutation on the assignment space of an unsatisfiable \CSOP{} is a symmetry. This makes Definition \ref{def_symmetry} a very general notion of symmetry, and in practice, only particular types of symmetry are considered. Some examples are \emph{value symmetry}, \emph{variable symmetry}, \emph{row symmetry} and \emph{column symmetry} \cite{compositional_sym_detection}. More often than not, symmetries are induced by permutations on the set of variables or the set of values of a \CSOP:

\begin{definition}
A \emph{variable symmetry} for a \CSOP{} $(V,D,C,O)$ is a symmetry $S_\pi: \alpha \mapsto \alpha \circ \pi$ induced by a permutation $\pi: V \rightarrow V$. A \emph{value symmetry} is a symmetry $S_\rho: \alpha \mapsto \rho \circ \alpha$ induced by a permutation $\rho: D \rightarrow D$.
\end{definition}

Note that Definition \ref{def_symmetry} does not require a symmetry to be invariant with respect to the objective function; a symmetry $S$ is \emph{invariant} for objective function $O$ iff $\forall \alpha: O(\alpha)=O(S(\alpha))$. When symmetry is used to reduce search time by eliminating symmetric parts of the search space through \emph{symmetry breaking}, the broken symmetry $S$ must preserve the implicit minimization constraint of a \CSOP{}. However, in a local search context, we will also investigate symmetries who are \emph{variant} (as in \emph{not invariant}) with respect to the objective function.

\begin{example}
\label{ex_chromatic}
The \emph{Chromatic Number Problem} (CNP) consists of identifying the minimum number of colors with which a graph can be colored such that each two adjacent nodes have a different color. We can model this problem as a \CSOP{} $(V,D,C,O)$ where each node is a variable in $V$, each possible color a value in $D$, the constraints $C$ state that two adjacent nodes in an input graph can not have the same color, and the objective function counts the number of colors used. %An assignment to the CNP represents a coloring of the nodes, regardless of the edges in the graph.

Any permutation $\rho$ of the domain $D$ induces a value symmetry $S_\rho$ for the CNP. Each such $S_\rho$ is invariant for the objective function.
\end{example}

\begin{example}
\label{ex_sym_in_TSP}
Using the TSP model from Example \ref{ex_TSP}, each permutation of the domain induces a value symmetry and each permutation of the variables induces a variable symmetry. Both of these symmetry classes are variant for the objective function.
\end{example}

\subsection{Symmetries induce a neighborhood}
\label{sec_induce_a_neighborhood}
Note that the 2-opt neighborhood of Example \ref{ex_TSP_neighborhood} is based on a particular set of permutations of the set of cities in the TSP-tour. It is striking that these permutations also induce symmetries for the TSP-problem. We formalize this connection between symmetry and neighborhood:

\begin{definition}
\label{def_induced_neighborhood}
Given a set of symmetries $\mathcal{S}$ for some \CSOP{}, the \emph{symmetry-induced neighborhood} $N_\mathcal{S}$ maps each satisfying assignment $\alpha$ to its image under $\mathcal{S}$. More formally, $N_\mathcal{S}: \alpha \mapsto \{S(\alpha)~|~S \in \mathcal{S}\}$.
\end{definition}

By Definition \ref{def_symmetry}, any satisfying assignment has only satisfying assignments as symmetry-induced neighbors, which ensures Definition \ref{def_induced_neighborhood} is a sound neighborhood definition.

Note however that Definition \ref{def_induced_neighborhood} requires some set of symmetries as input. Since a set of symmetries $\mathcal{S}$ forms a group $\Gamma_\mathcal{S}$ under functional composition, the number of possible symmetry sets to form neighborhoods with often is astronomical. In general, we have no definitive answer on what sets of symmetries one should use, but it seems plausible to use some small set $\mathcal{S}$ that generates the detected symmetry group $\Gamma_\mathcal{S}$. This way, each move possible under the induced neighborhood $N_{\Gamma_\mathcal{S}}$ can be simulated by a series of moves under $N_{\mathcal{S}}$, while $N_{\mathcal{S}}$ maps a satisfying assignment to a relatively small set of neighbors.

Using this notion of a symmetry-induced neighborhood, we can automatically compile a \CSOP{} specification to input for a local search algorithm solving the \CSOP{}. Recall that the only input required for many local search algorithms is some initial satisfying assignment $\alpha$ and a neighborhood $N$. The following steps generates these from only the problem specification:
\begin{enumerate}
\item Generate an initial satisfying assignment $\alpha$ using existing constraint programming technoloby.
\item Detect a symmetry group $\Gamma$ of the constraint optimization problem.
\item Use some set of symmetries $\mathcal{S} \subseteq \Gamma$ to construct a symmetry-induced neighborhood $N_\mathcal{S}$.
\end{enumerate}

\section{Automating Local Search in \idp}
\label{sec_FO_Theory}
%\jo{TODO: fix overloaded symbols and terms between CP and FO}

We implemented the automatic detection of symmetry-induced neighborhoods in the \CSOP{}-solving \idp system. In this section, we will sketch relevant details of the \idp system, as well as the type of symmetry and neighborhoods it detects. We also illustrate the detection by means of the TSP and CNP examples.

\subsection{\idp as a constraint solving system}
The \idp system is an experiment in constructing a \emph{knowledge base} system. The aim of a knowledge base system is to solve problems in a radically declarative way, where domain knowledge is specified once as a knowledge base, allowing the user to solve multiple domain problems without further modifications to the knowledge base. The specification language of \idp is \fodot -- an extension of typed classical first-order logic with aggregates, arithmetic and inductive definitions.

One of the problems the \idp system is capable of solving is a logical \emph{model optimization} problem (\LMOP{}). A \LMOP{} is the logical equivalent of a \CSOP{}, which we will show after the brief introduction of some logical concepts.

In \fodot, logical formulas are constructed using a \emph{vocabulary} \voc, which contains \emph{type} symbols, \emph{predicate} symbols and \emph{function} symbols. The type symbols specify sets of \emph{domain elements} present in the problem, while the predicate and function symbols specify respectively typed relations and typed functions. A \emph{structure} \struct over a vocabulary \voc is an association of actual sets, relations and functions to the symbols in \voc. More formally, we say a structure \struct \emph{interpretes} the symbols in \voc, while for a symbol $P \in \voc$, $P^\struct$ is called its \emph{interpretation}. Structures can be \emph{partial}, in which case some predicate or function symbols have no interpretation.
Given the symbols in a vocabulary, \emph{formula}'s and \emph{terms} can be constructed using logical connectives, quantifiers and other logical symbols. If a structure \struct ranges over the same vocabulary as a formula $\phi$ or a term $t$, then $\phi^\struct$ and $t^\struct$ are evaluations of $\phi$ and $t$: $\phi^\struct$ is either true or false, while $t^\struct$ maps to some domain element $d$ from $\struct$.\footnote{To be exact, $\phi$ and $t$ should also not contain any unquantified logical variables to have a proper evaluation in $\struct$.}
A \emph{theory} \theory over a vocabulary \voc is a set of formulas over \voc, and a structure \struct over \voc \emph{satisfies} \theory if for all $\phi \in \theory$, $\phi^\struct$ is true. In this case, we say that $\struct$ is a \emph{model} of $\theory$, or $\struct \models \theory$.

A \LMOP{} can be characterized as a quadruple $(\Sigma,\structt,\theory,t)$, where $\Sigma$ is the vocabulary, \structt a partial structure over $\Sigma$, \theory a theory over $\Sigma$ and $t$ a term over $\Sigma$ mapping to a numeric domain such as the natural numbers. A \LMOP{} $(\Sigma,\struct,\theory,t)$ then represents the task of finding
\begin{itemize}
\item a model $\struct \models \theory$,
\item such that \struct has the same interpretation as \structt for all types and interpreted symbols in \structt,
\item and \struct is minimal for $t$.\footnote{Again we assume the objective function is to be minimized.}
\end{itemize}
When relating this to a \CSOP{} $(V,D,C,O)$, \theory represents the constraints $C$, $t$ represents the objective function $O$, the uninterpreted function and predicate symbols in \structt represent the variables $V$, and the domain of values $D$ corresponds to the set of possible relations and functions that can be associated to the uninterpreted symbols, given the interpretation of the types of \voc in \structt. A structure corresponds to an assignment, a model to a satisfying assignment, and a minimal model to an optimal solution.

Instead of diving into the technical details of \fodot, we improve a reader's intuition by the following two examples:
\begin{example}
\label{ex_TSP_idp}
The \CSOP{} specification for TSP given in Example \ref{ex_TSP} can be realized as a \LMOP{} with the following vocabulary:
\begin{itemize}
\item type $City$
\item type $Index \subseteq \mathbb{N}$
\item function symbol $Distance: City\times City \rightarrow \mathbb{N}$
\item function symbol $Next: Index \rightarrow Index$
\item function symbol $Map: Index \rightarrow City$
\end{itemize}
For the TSP, every symbol but $Map$ will be interpreted by the partial structure, so $Map$ will function as our search variable.

The TSP constraints are specified by a theory with one formula:
\begin{align*}
\forall x: \forall y: (Index(x) \land Index(y) \land x\neq y) \Rightarrow Map(x)\neq Map(y).
\end{align*}
Which effectively posts an all-different constraint over the $Map$ function symbol, stating that two different index elements need to be mapped to two different cities. Each model satisfying this theory will thus have a TSP tour as interpretation for $Map$.

Next we need a minimization term:
\begin{align*}
\sum_{z} \{Distance(Map(z),Map(Next(z)))~|~z \in Index\}
\end{align*}
Which denotes the sum of all distances between cities mapped by subsequent indices, and as such denotes the total distance of a tour represented by the $Map$ function symbol.

Finally, the partial structure provides the necessary parameters to solve a TSP instance. For this, it contains
\begin{itemize}
\item a set of indices $\{0,\ldots,n-1\}$ and a set of cities $\{c_1,\ldots,c_n\}$ as interpretation for the types,
\item a function adhering to the signature of the $Distance$ symbol specified in the vocabulary,
\item a function adhering to the signature of the $Next$ symbol specified in the vocabulary, which for sensible TSP instances maps some index $x$ to $(x+1)\mod n$.
\end{itemize}

Given the above vocabulary, theory, minimization term and partial structure, \idp's \LMOP{} routine then searches for an interpretation to $Map$ that satisfies the constraints in the theory, and minimizes the objective function. This solves the TSP \CSOP{} described in Example \ref{ex_TSP}, as the inferred interpretation of $Map$ leads to an optimal TSP tour.
\end{example}

\begin{example}
\label{ex_chromatic_number}
The CNP from Example \ref{ex_chromatic} can be modelled in \fodot using the following vocabulary:
\begin{itemize}
\item type $Node$
\item type $Color$
\item predicate symbol $Edge \subseteq Node \times Node$
\item function symbol $Coloring: Node \rightarrow Color$
\end{itemize}

The constraint that two neighboring nodes must have a different color is stated in the following theory:
\begin{align*}
\forall x: \forall y: Edge(x,y) \Rightarrow Coloring(x)\neq Coloring(y).
\end{align*}
And the objective function simply counts the number of colors used:
\begin{align*}
\#\{z~|~\exists x: Coloring(x)=z\}
\end{align*}

Finally, a partial structure contains an input graph by interpreting $Edge$, and leaves the $Color$ symbol uninterpreted. Solving the \LMOP{} will lead to an interpretation for $Color$, representing a minimal graph coloring.
\end{example}

Algorithmically, \idp solves model optimization by a \emph{ground-and-solve} approach, where the  theory, structure and minimization term are \emph{grounded} to a set of low-level constraints in the \emph{extended conjunctive normal form} (ECNF) language. These ECNF constraints can be solved by \idp's custom-made \emph{lazy clause generation} constraint solver \minisatid \cite{ictai/DeCat13}. This grounding step is analogous to the \emph{flattening} of high-level MiniZinc specifications to FlatZinc constraints, to the point that \minisatid can also solve FlatZinc specifications.\footnote{\minisatid also participated in 2014's and 2015's FlatZinc competition.}

%Variables are specified in a logical \emph{vocabulary}, constraints in a logical \emph{theory}, and instance-specific information such as domains and parameters are specified in a logical \emph{structure}. Given these three, \idp follows a \emph{ground-and-solve} approach, turning the \fodot specification into a low-level \emph{extended conjunctive normal form} (ECNF) set of constraints, that can be solved by \minisatid, \idp's backend \emph{lazy clause generation} constraint solver\cite{ictai/DeCat13}. Converting \fodot specifications to ECNF is comparable to how MiniZinc specifications are \emph{flattened} to FlatZinc, to the point that \minisatid can also solve FlatZinc specifications\footnote{\minisatid even participated in last year's FlatZinc competition.}.

\subsection{Symmetry in \fodot}
Given that we defined a symmetry as a permutation of the assignment space preserving satisfaction to the constraints, a symmetry of a \LMOP{} $(\Sigma,\structt,\theory,t)$ corresponds to a permutation of the set of extensions of \structt such that for each extension $\struct$ of $\structt$ holds $\struct \models \theory \Leftrightarrow S(\struct) \models \theory$.

The symmetry type detected by \idp is \emph{domain element swap} (\idpsym{}) symmetry, which is a variant of the type of symmetry detected by the relational model finder Kodkod \cite{tacas/TorlakJ07}. Kodkod uses untyped first-order logic, and hence provides only one set of domain elements $Dom$ in its partial structure \structt. Kodkod's symmetry detection routine then partitions $Dom$ into subsets $Dom_i$ such that for any $Dom_i$ all swaps of domain elements $d_1,d_2 \in Dom_i$ induce a symmetry. \idp's \idpsym{} symmetry similarly exploits the given types as partition of the set of all domain elements, but it does not require all symbols ranging over the type to take part in the symmetry.

Let's provide a formal definition for clarification:

\begin{definition}
\label{def_\idpsym{}}
A \emph{domain element swap} (\idpsym{}) symmetry $S$ of a \LMOP{} $(\Sigma,\structt,\theory,t)$ is a symmetry characterized by a triple $(a,b,\sigma)$ such that $a,b$ are two domain elements from the same type interpretation in $\structt$, and $\sigma$ is a subset of predicate and function symbols from $\Sigma$ that are uninterpreted in \structt. If we take $\pi_{ab}$ to be the permutation of domain elements that swaps $a$ with $b$ and leaves all other domain elements invariant, then $S$ maps each extension \struct of \structt to $S(\struct)$ in such a way that for predicate symbols $P \in \sigma$:
\begin{align*}
(d_1,\dots, d_n) \in P^{\struct} \Leftrightarrow (\pi_{ab}(d_1),\dots, \pi_{ab}(d_n)) \in P^{S(\struct)}
\end{align*}
and for function symbols $f \in \sigma$:
\begin{align*}
d_0=f^{\struct}(d_1,\dots, d_n) \Leftrightarrow \pi_{ab}(d_0)=f^{\struct}(\pi_{ab}(d_1),\dots, \pi_{ab}(d_n))
\end{align*}
while the interpretation of predicate symbols $Q \not \in \sigma$ and function symbols $g \not \in \sigma$ is left untouched:
\begin{align*}
Q^{\struct}=Q^{S(\struct)} \text{ and } g^{\struct}=g^{S(\struct)}
\end{align*}
\end{definition}

%Note that a \idpsym{} symmetry $S$ for a \LMOP{} potentially is a combination of both variable and value symmetry in the \CSOP{} sense: if any output position for some function symbols occurs in $S$'s argument list, it permutes the \CSOP{}'s corresponding values.

The following two examples illustrate \idpsym{} symmetry:

\begin{example}
\label{ex_TSP_sym}
The TSP \LMOP{} from Example \ref{ex_TSP_idp} exhibits two classes of \idpsym{} symmetry:
\begin{itemize}
\item \idpsym{} symmetries characterized by $(i,j,\{Map\})$ for $i,j \in [0 \ldots  n-1]$. These symmetries swap two indices $i$ and $j$ in the interpretation of $Map$, and as such are equivalent to the TSP variable symmetry of Example \ref{ex_sym_in_TSP}.
\item \idpsym{} symmetries characterized by $(c_i,c_j,\{Map\})$ for $i,j \in [1\ldots n]$. These symmetries swap two cities $c_i$ and $c_j$ in the interpretation of $Map$, and as such are equivalent to the TSP value symmetry of Example \ref{ex_sym_in_TSP}.
\end{itemize}
\end{example}

\begin{example}
\label{ex_chromatic_number_sym}
The chromatic number \LMOP{} from Example \ref{ex_chromatic_number} exhibits the following class of \idpsym{} symmetry:
\begin{itemize}
\item \idpsym{} symmetries characterized by $(c,c',\{Coloring\})$ for two domain elements $c, c'$ in $Color$'s interpretation. These symmetries swap two colors $c$ and $c'$ in the interpretation of $Coloring$, and as such are equivalent to the chromatic number value symmetry of Example \ref{ex_chromatic_number}.
\end{itemize}
\end{example}

\subsection{Symmetry-induced neighborhood detection in \idp}
\label{sec_neighb_detection_idp}

As mentioned at the end of Section \ref{cp_ls}, we can convert a \CSOP{} specification to input for a local search algorithm by use of a symmetry-induced neighborhood. The previous section explained the type of symmetry the \idp system can detect for a \LMOP{}%\footnote{The actual \idpsym{} symmetry detection algorithm of \idp is out of the scope of this paper.}
, so all that is left to do is figure out which of these symmetries induce good neighborhoods.

We put forward that symmetries that are invariant for the objective function are bad candidates for neighborhood generation. Often such symmetries constitute a simple renaming of variables, values or domain elements, and as such can not transform a satisfying assignment into a reasonably different one. Of course, sometimes a move in a local search algorithm transforms the current satisfying assignment into one with the same objective value, but having a neighborhood that \emph{only} leads to such moves seems like a waste of resources. We shortly investigate some properties of \idpsym{} symmetry present in a \LMOP{} problem with regard to the invariance of the objective function.

Firstly, a sufficient condition for a \idpsym{} symmetry $(a,b,\sigma)$ to leave the minimization term $t$ %in a \LMOP{} $(\Sigma,\structt,\theory,t)$ 
invariant is that $\sigma$ does not contain any predicate or function symbols occurring in $t$. As such, we can instruct \idp's neighborhood detection algorithm to only investigate symmetries ranging over some uninterpreted symbol in $t$. For the TSP problem specification from Example \ref{ex_TSP_idp} the optimization term contains the uninterpreted symbol $Map$, which occurs in both symbol lists of the TSP \idpsym{} symmetries given in Example \ref{ex_TSP_sym}. As a result, it is possible that both symmetry classes are not invariant for the minimization term, which upon further inspection is the case.

However, ranging over an uninterpreted symbol in the minimization term is not a necessary condition for a \idpsym{} symmetry to be invariant for the minimization term, as shown by the CNP. The CNP \LMOP{} minimization term (see Example \ref{ex_chromatic_number}) contains the function symbol $Coloring$, which also occurs in the list of symbols of the \idpsym{} symmetry given in Example \ref{ex_chromatic_number_sym}. However, swapping two colors in a graph coloring is invariant for the number of colors used. \idp's symmetry detection scheme utilizes more refined mechanisms to detect whether terms and formula's are invariant under a certain \idpsym{} symmetry, which we will use in the experiments, but which will not be explained in further detail here.

With the above points in mind, we can devise a simple symmetry-induced neighborhood detection scheme for a \LMOP{} $(\Sigma,\structt,\theory,t)$ in \idp:
\begin{enumerate}
\item Identify the predicate and function symbols occurring in $t$ but uninterpreted in \structt.
\item Detect \idpsym{} symmetry over these symbols.
\item Ignore any \idpsym{} symmetry invariant for $t$.
\item Convert the remaining \idpsym{} symmetries to neighborhoods by Definition \ref{def_induced_neighborhood}.
\end{enumerate}

The performance critical part of this algorithm is the symmetry detection step, whose efficiency in turn depends on the granularity of the symmetry detected. For \idp, symmetry detection takes at most $O(n^2)$ time, with $n$ the total number of domain elements in \structt, since worst-case it generates all pairs of domain elements $a,b$ to check for \idpsym{} symmetries $(a,b,\sigma)$. The list of symbols $\sigma$ is derivable in linear time from \theory. So for \idpsym{} symmetries, the neighborhood detection mechanism is tractable.

The only question that remains is what (small) set of symmetries should be used to induce neighborhoods, as was mentioned at the end of Section \ref{sec_induce_a_neighborhood}. 
Note that \idpsym{} symmetries represent swaps of domain elements, which can be composed to form other symmetries based on other permutations of domain elements. Moreover, any permutation over a set of domain elements can be obtained by a composition of swaps of domain elements. As a result, detecting a set $\mathcal{S}$ of \idpsym{} symmetries entails detecting a group of symmetries $\Gamma_\mathcal{S}$ that represents the interchangeability of subsets of domain elements (those that can be pairwise swapped).

The size of $\Gamma_\mathcal{S}$ is factorial in the size of $\mathcal{S}$, so using all symmetries in $\Gamma_\mathcal{S}$ as neighborhood inducing symmetries seems an infeasible option. Instead, we restrict the set of neighborhood inducing symmetries to the set of all possible swaps of domain elements, in casu $\mathcal{S}$. This has two advantages: firstly it limits the amount of neighborhood inducing symmetries to $O(n^2)$ with $n$ the number of swappable domain elements. Secondly, $\mathcal{S}$ \emph{generates} $\Gamma_\mathcal{S}$, meaning that any model $S(\struct)$ that can be reached by $S \in \Gamma_\mathcal{S}$ from model $\struct$ can also be reached by a composition of some series of $S' \in \mathcal{S}$.

On the other hand, it is possible to construct a set of symmetries $\mathcal{S}'$ that generates $\Gamma_\mathcal{S}$ but which is $O(n)$ in size. $\mathcal{S}'$ then consists of swaps of subsequent domain elements $d_i,d_{i+1}$ according to some chosen total order on the domain elements. Even though it would lead to smaller neighborhoods, it also skews any local search algorithm according to the chosen order, putting a possibly unwarranted bias on the direction of the search over the solution space. For this reason, we stick with the quadratic set of symmetries $\mathcal{S}$ to induce a neighborhood.

\section{Experiments}
\label{sec_experiments}
We implemented the neighborhood detection scheme described in the previous section in \idp, and in this section we experimentally investigate which neighborhoods were detected for a series of well-known constraint optimization problems. The \fodot specifications for each of these problems are available online at \url{adams.cs.
kuleuven.be/idp/localsearch.html}\footnote{Click ``File'', then ''Local Search in \idp''}, where an interested reader can run \idp with one click and see the neighborhood detection mechanism in action.

\subsection{Traveling Salesman Problem: on robustness}
Let us first investigate the TSP problem, since this was our running example. As mentioned in Section \ref{sec_FO_Theory}, both the city-swapping and index-swapping symmetries are variant for the minimization term, resulting in a neighborhood swapping cities and indices.

However, there exist other reasonable \fodot specifications of TSP other than the one in Example \ref{ex_TSP_idp}. For instance, a user could use the following vocabulary, theory and minimization term specifying the TSP:\\

Vocabulary:
\begin{itemize}
\item type $City$
\item function symbol $Distance:City\times City \rightarrow \mathbb{N}$
\item predicate symbol $Following \subseteq City \times City$
\item predicate symbol $Reachable \subseteq City$
\item constant $Start:~\rightarrow City$
\end{itemize}

Theory:
\begin{align*}
&\forall x: \exists1 y: Following(x,y).\\
&\forall y: \exists1 x: Following(x,y).\\
&\forall x: Reachable(x).\\
&\{\forall x: Reachable(x) \leftarrow x=Start \vee (\exists y: Reachable(y) \wedge Following(y,x)).\}
\end{align*}
The constraint between curly brackets is an \emph{inductive definition} \cite{tocl/DeneckerT08}, constraining the $Reachable$ predicate to only be true for cities reachable from a $Start$ city using the $Following$ relation. By next stating that all cities must be reachable, we effectively posted a subtour elimination constraint. \\

Minimization term:
\begin{align*}
\sum_{x,y} \{Distance(x,y)|Following(x,y)\}
\end{align*}

When executing the described neighborhood detection algorithm, we establish that $Following$ is an uninterpreted predicate symbol occurring in the objective function, and that \idpsym{} symmetries swapping cities over this symbol exist. Also, these symmetries are variant for the objective function, so they lead to a city-swapping neighborhood.

This alternative specification experiment shows that the proposed neighborhood detection method exhibits robustness: different specifications of the same problem still lead to comparable neighborhoods. This of course only holds as long as symmetry properties of different specifications are similar.

\subsection{Shortest Path Problem: a succesful problem}
This problem consists of finding the shortest path between a start and end node in a weighted graph. Its specification in \fodot is very similar to the TSP specification of the previous subsection, and centers on finding a minimal interpretation to some $Following/2$ predicate. \idp's neighborhood detection mechanism indeed detected that all two cities except the start and end city were interchangeable. The resulting symmetries were variant for the objective function, leading to a large set of induced neighborhoods. Therefor, we judge \idp's neighborhood detection algorithm to be succesful on the shortest path problem.

\subsection{Max Clique: relaxing constraints?}
The task of a max clique problem is to identify the largest clique in a graph. We modelled this problem in such a way that each satisfying assignment represented a set of nodes forming a clique in the input graph. The only domain elements in this problem are the nodes of the graph, and for typical graphs nodes are not swappable. Hence, \idp could not detect any symmetry, and no induced local search neighborhoods were detected by our algorithm. This makes sense, since there is no obvious way of transforming cliques of a graph in one move to some other clique in the graph. 

However, we can imagine a local search algorithm for this problem that iteratively removes nodes from and to a node set representing a potential clique. This effectively \emph{relaxes} the clique constraint, making any set of nodes a satisfying assignment, regardless of whether it forms a clique in the input graph. The lesson we take from this problem is that our neighborhood detection scheme is not yet able to detect constraints that can be relaxed to allow for a larger potential neighborhood.

\subsection{Chromatic Number Problem: avoiding a useless neighborhood}
As mentioned in Example \ref{ex_chromatic}, the CNP consists of finding the lowest amount of colors with which a graph can be colored so that no two adjacent nodes have the same color. This problem does exhibit symmetry in the sense that all colors to color the nodes with are swappable. However, as mentioned in Section \ref{sec_neighb_detection_idp}, this symmetry is invariant for the objective function, and as a result, no symmetry-induced neighborhood was detected. This is a positive result, since globally swapping colors does not lead to an effective local search neighborhood.

However, swapping colors for each node separately might lead to a reasonable neighborhood, which would again relax the constraint that two adjacent nodes must have a different color. This observation is similar to the one made in the Max Clique section.

\subsection{Knapsack Problem: unexpected symmetry leads to unexpected neighborhood}
The knapsack problem consists of filling up some abstract knapsack with objects, such that the volume of the objects fits the knapsack, but the value of the objects is maximal. Since swapping any two objects in and out of the knapsack might violate the volume constraint, we expected the neighborhood detection algorithm to not detect any symmetry, and thus no neighborhood.

However, \idp's symmetry detection was sufficiently fine-grained to identify that for some instances, some objects had the same volume but a different value. These objects could safely be swapped in and out of the knapsack, leading to an unexpected neighborhood where for a given knapsack, small variations on the filling of the knapsack could be explored.

\subsection{Assignment Problem: human neighborhoods}
The last problem we investigate is the assignment problem, where a bijection between a set of agents and a set of tasks must be found that minimizes the cost of assigning a certain task to a certain agent. As expected, \idp detected that swaps of agents and tasks were not a priori invariant for the objective function. As a result, these symmetries induced a local search neighborhood, which arguably would be the same neighborhood a human algorithmician would devise. This is an optimal neighborhood detection result!

\subsection{Experimental conclusions}
Testing the symmetry-based automated neighborhood detection algorithm of \idp on the above problems yielded interesting insights. Firstly, the approach is robust in changes to the specification as long as the symmetry properties are not disturbed. Secondly, taking invariantness of the minimization term into account allows to avoid detecting useless neighborhoods. Thirdly, for some problems, the symmetry-induced neighborhood is the same as the one a human would devise. Fourthly, to detect more neighborhoods, it might be needed to relax constraints. And finally, sometimes the automated neighborhood detection algorithm might find neighborhoods where a human did not expect them.

On the whole, we evaluate the experiment to have returned positive results, supporting the viability of symmetry-induced neighborhood detection.

\section{Related Work}
In the previous sections we described how any constraint programming system and how \idp in particular can exploit symmetry properties of problems to derive local search neighborhoods for those problems. Of course, we are not the first to try to link local search to constraint programming. 
A well-known example is the Comet system, which allows a user to easily specify neighborhood based local search algorithms in a constraint-centered way \cite{cp/MichelH05,bookHentenryck2005}. However, Comet did not provide any automatic neighborhood detection algorithm.

Stochastic SAT solvers such as WalkSAT \cite{dimacs/SelmanKC95} also allow a constraint programming problem to be solved by local search. In WalkSAT, every assignment to the boolean variables is a satisfying assignment, and neighborhoods are defined in terms of ``flips'' on the truth value of a boolean variable. In our view, this is an extreme approach, since all original constraints can be violated by any move. Note that in a context where all constraints are relaxed, any permutation on the solution space is a symmetry which can be used for symmetry-induced neighborhoods. However, these neighborhoods are not the kind a human programmer would devise for most combinatorial optimization problems.

To our knowledge, the only system that allows automatic derivation of local search neighborhoods from an input specification is LocalSolver \cite{localsolver}. LocalSolver allows a user to write down constraints in a mathematical modeling language centered on boolean variables, and uses flips on those variables as well as feasibility preserving moves based on ``ejection chains applied to the hypergraph induced by boolean variables and constraints''. We conjecture that these feasibility preserving moves can be seen as symmetries of the problem, but we need further investigation to confirm this. One weakness of LocalSolver is that the initial solution is found by a basic randomized greedy algorithm, and that it is not designed for solving hardly-constrainted optimization problems. This is opposed to the constraint-based local search approach described in this paper, which can always fall back on the solving capabilities of a state-of-the-art constraint programming engine.

To put our work in a broader perspective, it is worth mentioning that much research has been done on the relationship between symmetry \emph{breaking} and local search (e.g. \cite{symmetry_breaking_in_LS}). An important result is that adding symmetry breaking constraints has a negative impact on the efficiency of local search algorithms. In our work, we do not break any symmetry, but rather exploit symmetries to traverse the local search space. Most importantly, the symmetries we use are not symmetries of the whole optimization problem, since they are variant for the objective function. As such, they cannot be broken, since this would risk removing an optimal solution from the search space. In this light, symmetry breaking for optimization problems and exploiting symmetry properties to construct local search neighborhoods are two orthogonal uses of symmetry.

On the other hand, neighborhood inducing symmetries share the fact that they are variant for the objective function with \emph{dominance relations} used in \emph{dominance breaking}. Dominance breaking is a generalization of symmetry breaking for complete search algorithms where extra constraints remove \emph{dominated} solutions with a worse objective value from the search space \cite{Chu:2012:GMI:2405292.2405298}. These dominance relations seem to also induce neighborhoods much in the same way symmetry does and can be detected automatically \cite{conf/ijcai/MearsB15}, but it is not clear if dominance relations of a \CSOP{} are fundamentally different to symmetries that are variant for the objective function.

Finally, in recent years the metaheuristic community is actively investigating how to formalize and automate local search algorithms \cite{mic/metaheuristicagenda}. Our work can be seen as an effort in that direction, opening up unexplored research avenues by linking local search neighborhoods to symmetry properties of problems.

\section{Conclusion and future work}
In this paper, we propose a link between local search neighborhoods and symmetry. To our knowledge, this is the first time such a link is established. We used this link to design and implement a transformation of a \CSOP{} specification, in the form of a \LMOP{} specified in \fodot, to input for an automated local search algorithm. We conducted an experimental investigation of the symmetry-induced neighborhoods detected by our implementation, which supports the viability of this technique.

As future work, it remains to be seen how well symmetry-based neighborhood detection algorithms perform on larger, more complex problems, with potentially more symmetry breaking constraints. Allowing the relaxation of certain constraints might prove crucial in detecting sufficiently large neighborhoods. Also, the link with techniques for dominance breaking is definitely worth investigating.

Secondly, it would be interesting to couple the detected neighborhoods and input solutions to some local search engine, and experimentally verify their performance. The challenge here lies in being able to quickly move from one satisfying assignment to another, and to incrementally update the value of an objective function. The \idp system is currently unoptimized in this regard.

Thirdly, the problem of deciding which subset of detected symmetries are used to induce neighborhoods can also be tackled by a \emph{hyperheuristic} algorithm. In general, a hyperheuristic aims to find the best combination of neighborhoods by deciding which neighborhood to select for the next iterations of local search and evaluating the resulting executed moves. Providing information on the whole symmetry group to the hyperheuristic algorithm instead of only a subset of symmetry-induced neighborhoods might allow it to derive more fitting combinations of neighborhoods.

Lastly, for some problems, there exist complex neighborhoods involving smart perturbation and repair steps. Detecting these still seems a hard challenge.

\section*{Acknowledgements}
Thanks go to Ingmar Dasseville for configuring the \idp web interface with our local search examples, and to Bart Bogaerts and the anonymous reviewers for helpful comments that improved this paper.

%The bibliography, done here without a bib file
%This is the old BibTeX style for use with llncs.cls
\bibliographystyle{splncs}

%Alternative bibliography styles:
%the following does the same as above except with alphabetic sorting
%\bibliographystyle{splncs_srt}
%the following is the current LNCS BibTex with alphabetic sorting
%\bibliographystyle{splncs03}
%If you want to use a different BibTex style include [oribibl] in the document class line

\bibliography{krrlib}

\end{document}